\documentclass[11pt,draftcls, onecolumn]{IEEEtran}

\usepackage{epsfig}
\usepackage{changebar}
\usepackage{amssymb}
\usepackage{amsfonts}
\usepackage{amsmath}
\usepackage{mathrsfs}
\usepackage{amstext}
\usepackage{wasysym}
\usepackage{epstopdf}
\usepackage{graphicx}

\title{Classification via Incoherent Subspaces}

\author{Karin Schnass, \and Pierre Vandergheynst, {\em Senior Member, IEEE}
\thanks{Karin Schnass is with the Johann Radon Institute for Computational and Applied Mathematics (RICAM), Altenbergerstrasse~54, 4040 Linz, Austria,
	E-mail: karin.schnass@oeaw.ac.at }
\thanks{Pierre Vandergheynst is with the Signal Processing Laboratory~2,
Swiss Federal Institute of Technology (EPFL), Station~7, 1015 Lausanne, Switzerland,
E-mail: pierre.vandergheynst@epfl.ch}
\thanks{This work was partly supported by NSF grant 200021-117884/1}
}

\newcommand\eps{{\varepsilon}}

\newcommand\sig{{y}}
\newcommand\signew{{\sig_{new}}}
\newcommand\class{{Y}}
\newcommand\X{{\class}}

\newcommand\coef{x}

\newcommand\feat{{f}}
\newcommand\featnew{{\feat_{new}}}

\newcommand\Feat{{F}}
\newcommand\Fn{{F}}
\newcommand\fn{{f}}

\newcommand\sign{{\operatorname{sign}}}
\newcommand\prox{{\operatorname{prox}}}
\DeclareMathOperator*{\argmin}{argmin}
\DeclareMathOperator*{\argmax}{argmax}
\newcommand\ip[2]{\langle #1, #2\rangle}

\newcommand\nsig{{N}}
\newcommand\ddim{{d}}

\newcommand\nclass{{c}}
\newcommand\n{{n}}
\newcommand\nf{{s}}

\newcommand{\R}{{\mathbb{R}}}

\newcommand{\ie}{{i.e. }}

\newtheorem{Theorem}{Theorem}

\begin{document}

\date{}

\maketitle

\begin{abstract}
This article presents a new classification framework that can extract individual features per class. The scheme is based on a model of incoherent subspaces, each one associated to one class, and a model on how the elements in a class are represented in this subspace. After the theoretical analysis an alternate projection algorithm to find such a collection is developed. The classification performance and speed of the proposed method is tested on the AR and YaleB databases and compared to that of Fisher's LDA and a recent approach based on on $\ell_1$ minimisation. Finally connections of the presented scheme to already existing work are discussed and possible ways of extensions are pointed out.\\
\end{abstract}

%EDICS: SPC-CODC 

%%%%%%%%%%%%%%%%%%%%%%%%
\section{Introduction}
%%%%%%%%%%%%%%%%%%%%%%%%
A general approach in classification is to select features of the signal at hand and to get a decision by comparing them to the equivalent features of already labelled signals with a simple classifier like nearest neighbour, e.g. \cite{brpo93}, or nearest subspace, cp \cite{lehokr05}. This of course raises the question which features to take. For face recognition, which is the example we will use here, some classic and simple, because linear, features are Eigen, \cite{eigenfaces}, Fisher, \cite{lda}, or Laplace features, \cite{laplacefaces}. 
However, as these classifiers are very simple and the features not adjusted to them, their performance is somehow disappointing, and researchers turned to the development of more complicated nonlinear features and kernel methods, \cite{liu06,phsc07}.\\
Here we start from the point of view that the potential of linear methods and simple classifiers is not exhausted. In order to achieve better results, we propose to give up the uniformity of features over classes and mix the feature selection with the classifier. To motivate the idea of class specific features let us have a look at classical nearest neighbour (NN) and nearest subspace (NS) classification using linearly selected features and give it a new interpretation.\\
Assume we have $\nsig$ already labelled training signals $\sig\in\R^\ddim$ belonging to $\nclass$ classes, where each class $i$ contains $\n_i$ elements, \ie $\sum_i \n_i=\nsig$. We denote the $j$-th signal in class $i$ as $\sig_i^j$, $i=1\ldots \nclass, \, j=1\ldots \n_i$. For each class $i$ we collect all its training signals as columns in the $\ddim\times \n_i$ class matrix $\class_i$, \ie $\class_i=(\sig_i^1\ldots \sig_i^{\n_i})$, and these class matrices in turn are combined into a big $\ddim \times \nsig$ data matrix $\class=(\class_1\ldots \class_\nclass) = (\sig^1_1 \ldots \sig^{n_1}_1 \ldots \sig^1_\nclass \ldots \sig^{n_\nclass}_\nclass)$. Given a new signal $\signew$ the goal is to decide which class it belongs to with the help of the already labelled training signals.\\
The classical first step is to select relevant features $\featnew$ from $\signew$ via a linear transform $A$, where $A$ is a $d\times d$ matrix of rank $r \leq d$.
\begin{align}
\mbox{\bf Feature Selection: } \featnew = A \signew.
\end{align}
The exact shape of the transform is determined by the training signals and their labels. For instance for Fisher's LDA $A$ is chosen as the orthogonal projection that maximises the ratio of between-class scatter to that of within-class scatter, \cite{lda}.\\
In the second step these features are compared to the features $\feat_i^j: =A \sig_i^j$ of the training signals $y_i^j$. In case of the nearest neighbour classifier this means that the new signal will get the label of the training signal which has features that maximally correlate with the features of the new signal, ie.
\begin{align}
\mbox{\bf NN Labelling: } i_{new}=\argmax_i \max_j |\ip{\feat_i^j}{\featnew}|.
\end{align}
If by analogy we define the feature matrix for a class $i$ as $\Feat_i=A Y_i$ we can rewrite the expression, for which we are seeking the maximal argument, and combine the feature selection with the labelling step,
%\begin{align}
%\argmax_i \max_j |\ip{\feat_i^j}{\featnew}|&=\argmax_i \| F_i^\star \featnew\|_\infty \notag \\
%&=\argmax_i \| Y_i^\star A^\star A \signew\|_\infty 
%\end{align}
\begin{align}
\max_j |\ip{\feat_i^j}{\featnew}|&=\max_j |\ip{A\sig_i^j}{A \signew}|\notag \\
&= \max_j |\ip{A^\star A\sig_i^j}{\signew}| \notag \\
&= \| (A^\star A\class_i)^\star\, \signew\|_\infty,
\end{align}
where the matrix $M^\star$ denotes the transpose of $M$, the $p$-norm of a vector is defined by $\|v\|_p:=(\sum_k v(k)^p)^{1/p}$ for $1\leq p < \infty$ and $\|v\|_\infty:=\max_k |v(k)|$ and the $qp$-norm of a matrix by $\|M\|_{q,p}=\max_{\|v\|_q=1} \|Mv\|_p$. Thus another way of looking at the classification procedure is to say that for every class we have a set of sensing signals $A^\star A\sig_i^j$ and the new signal belongs to the class which has the sensing signal closest to it. From this point of view we also see that the scheme will work stably only if two conditions are fullfilled. Every new signal is well represented by one vector in its class, \ie a lot of its energy is captured by the projection on one vector, and no two sensing vectors from different classes are the same or close to each other, \ie
\begin{align}
\max_{i\neq k, j,l} |\ip{A^\star A\sig_i^j}{A^\star A\sig_k^l}| = \|(A^\star A\class_i)^\star (A^\star A\class_k)\|_{1,\infty}\leq \mu .
\end{align}
Let us do the same analysis for the nearest subspace classifier. Again the features of the new signal are compared to those of the training signals. For each class the features of the training signals in it span a subspace and the new signal will get the label of the class for which the orthogonal projection of the features of the new signal on the corresponding subspace has the highest energy. Let $Q_i$ be an orthonormal system\footnote{$Q_i$ can for instance be found via a (reduced) qr-decomposition of $AY_i$} spanning the subspace for class $i$, then \ie
\begin{align}
\mbox{\bf NS Labelling: } i_{new}=\argmax_i \|Q_i^\star \featnew\|_2.
\end{align}
Again we can combine the feature selection with the labelling by manipulating the expression, we want to maximise,
\begin{align}
\|Q_i^\star \featnew\|_2 = \| (A^\star Q_i)^\star \signew \|_2 . \notag
\end{align}
If we compare to NN classification we see that again for every class we get a set of sensing signals, the columns of the matrix $A^\star Q_i$, and that the new signal belongs to the class for which the sensing signals can take out the most energy (or for which the biorthogonal system to $(AQ_i)^\star$ provides the best representation). Again this leads to two conditions for the classification to work, which are however more complex. First every new signal should be comparatively well represented in the biorthogonal system $(Q_i^\star A)^\dagger$ determined by its class and second no signal which is in the span of the sensing signals of one class should be well representable in the biorthogonal system of another class. 
\begin{align}
\max_{i\neq k} \max_{\|x\|=1}\|(A^\star Q_i)^\dagger A^\star Q_j x\|_2 = \|(A^\star Q_i)^\dagger A^\star Q_j x\|_{2,2}  \leq \mu .
\end{align}
Summarising our findings for both nearest neighbour and nearest subspace classification we see that in both cases for every class we have a set of sensing signals or a subspace defined by the feature selection transform. There is a model how signals from this class are represented by this subspace, which implicitly determines which norm is used for the classification, $\|\cdot\|_\infty$ for NN, $\|\cdot\|_2$ for NS, and at the same requires that the interaction of subspaces measured by a corresponding matrix norm is small, \ie that they are incoherent.\\
The classification scheme presented in this paper is based on the following idea. We give up the restriction that the subspaces associated to each class are generated canonically as a function of the feature selection transform and the training samples, \ie $A^\star A Y_i$ in the case of nearest neighbour classification, but generate them individually. This idea can also be motivated using the example of face recognition, to which we will apply our scheme later. Uniform feature extraction would mean realising that in general the most relevant parts of a face are the regions of the eyes, nose and mouth. Thus in order to classify a person we would focus on the eyes, nose and mouth regions while ignoring the hairstyle and comparing them to the eyes, nose and mouth regions of all the candidates. While this makes sense in general it will fail as soon as the set of candidates contains identical twins which can only be distinguished by the birth mark one has on his cheek. So while for most people the cheek is not a very distinguishing feature for the twins it is and it would be better to remember for them the cheek instead of for instance the nose. Even without the extreme example of the identical twins individual features are natural considering that the people we meet every day all have eyes, mouths and noses but not all of them have distinguishing eyes, mouths and noses. Instead they may have distinguishing birthmarks, scars, chins, etc. and a representation using these features will characterise them well but nobody else.\\
In the next section we will introduce the mathematical framework on which we base our classification scheme. It consists of a model of subspaces associated to each class and a model of how the elements in this class are represented in this subspace, which together lead to a natural choice of the norm we have to use for the classification and an incoherence requirement on the subspaces. In Section~\ref{sec:find_sensing} we will develop a comparatively simple algorithm to learn these subspaces from the training signal, which we will use to classify faces in Section~\ref{sec:testing}. In the last section we summarise our findings, point out connections to related approaches and outline possibilities for future work.

%%%%%%%%%%%%
\section{Class Model}
%%%%%%%%%%%%
The most general model for the subspaces we can think of is to ascribe to every class $i$ a set of $\nf_i$ vectors $\fn_i^j$, $j=1\ldots s_i$, which are collected as columns in the matrix $\Fn_i= (\fn_i^1\ldots \fn_i^{s_i})$. These correspond to the features that characterise elements of this class, so every element $\sig_i$ in class $i$ can be written as a combination of these class specific features with coefficients $\coef_i$ and some residual $r_i$, orthogonal to the feature span,
\begin{align}
\sig_i= \Fn_i \coef_i + r_i, \hspace{1cm} r_i^k \perp sp(F_i).
\end{align}
The condition that features of a class well characterise the elements in it translates into a property of the coefficients $\coef_i$, \ie when measuring their strength in some norm it is higher than the strength of the coefficients we would obtain trying to represent the element by features of the wrong class.
Since without further restrictions on the set of features per class it is not straightforward to calculate the coefficients of the best representation of signal in a class, we will sacrifice generality for simplicity and for the rest of the analysis assume that for every class we have the same number of features $\nf$ and that they form an orthonormal system, \ie $ \Fn_i^\star \Fn_i = I_s$. We will point out how to deal with the more general situation in the last section. Given that the features of each class form an orthonormal system we can easily calculate the coefficients of the best representation of a general signal $y$ in class  $i$ as $\Fn_i^\star \sig$. The question is now how should we measure these coefficients in order to correctly classify our images, \ie which norm $\|\cdot\|$ should we choose such that for all $\sig_i$ in class $i$ we have
\begin{align}
\frac{\| \Fn_j^\star \sig_i \|}{\| \Fn_i^\star \sig_i \|} < 1,\, \forall j \neq i.
\end{align}
To answer this question we will introduce three models on the coefficients, each leading to a certain p-norm as optimal measure. 

%%%%%%%%%%%%%%%%%
\subsection{Sparse Coefficients}
%%%%%%%%%%%%%%%%%
Assume that all signals we want to classify can be well represented by one element of one class, \ie
\begin{align}
\sig_i= \Fn_i \coef_i + r_i \: \mbox{ with } \: \|\coef_i\|_0 = 1,
\end{align}
where $\|\cdot\|_0$ counts the number of non-zeros entries.
An example for this situation would be trying to sort pictures of monkeys, snails, cucumbers and broccoli into animal and vegetable pictures. Even though monkeys and snails are both animals their shapes are very different, meaning that we can think of them as orthogonal, and the same goes for the shapes of cucumbers and broccoli in the other class.
Let $x$ be the absolute value of the only non-zero component of the coefficients $\coef_i$. We immediately see that whatever p-norm we apply using the correct class the response is always equal to $x$, $\| \Fn_i^\star \sig_i \|_p=\|\coef_i\|_p = x$. Therefore to find out which p-norm is best we will use a trick that involves estimating the ratio we need to be smaller than 1 for successful classification with the triangular equation and a matrix norm bound. So for general $1 \leq p,q \leq \infty$ we get,
\begin{align}\label{pq-bound}
\frac{\|\Fn_j^\star \sig_i\|_p}{\|\Fn^\star_i\sig_i\|_p} &\leq \|\Fn_j^\star \Fn_i \|_{q,p} \frac{\|\coef_i\|_q}{\|\coef_i\|_p} +  \frac{\| \Fn_j^\star r_i\|_p}{\| \coef_i\|_p}.
\end{align}
In the special case where the coefficients are sparse and thus $\|\coef\|_q=\|\coef\|_p,  \forall p,q$, this means that
\begin{align}
\frac{\|\Fn_j^\star \sig_i\|_p}{\|\Fn^\star_i\sig_i\|_p} &\leq \|\Fn_j^\star \Fn_i \|_{q,p} +  \frac{\| \Fn_j^\star r_i\|_p}{x}.\notag
\end{align}
The smallest $qp$ norm of a matrix is obtained when $p=\infty$ and $q=\infty$. Then it corresponds to the maximal absolute entry of the matrix $\Fn_j^\star \Fn_i$, \ie the maximal absolute correlation between to features from different classes. Since in that case also the response from the residual $\| \Fn_j^\star r_i\|_p$ is minimal we get the best bound choosing the $\infty$-norm for the classification. 
Summarising our findings we see that in case of a sparse model on the coefficients, the $\infty$ norm is optimal and that the incoherence requirement we get for classification to work stably is that no two features from two different classes are too similar, but it does not matter if a feature is moderately close to all features in a different class or even representable by them. Thinking to the example of the animal vs. vegetable pictures this means that even though you can approximate the shape of a snail combining the shape of the cucumber and the broccoli, classification using the $\infty$ norm will work well because no animal shape alone closely resembles a vegetable shape and vice versa.

%%%%%%%%%%%%%%%
\subsection{Flat Coefficients}
\begin{figure}[htb]
%\centering
\begin{tabular}{ccccc}
 \includegraphics[width=3cm]{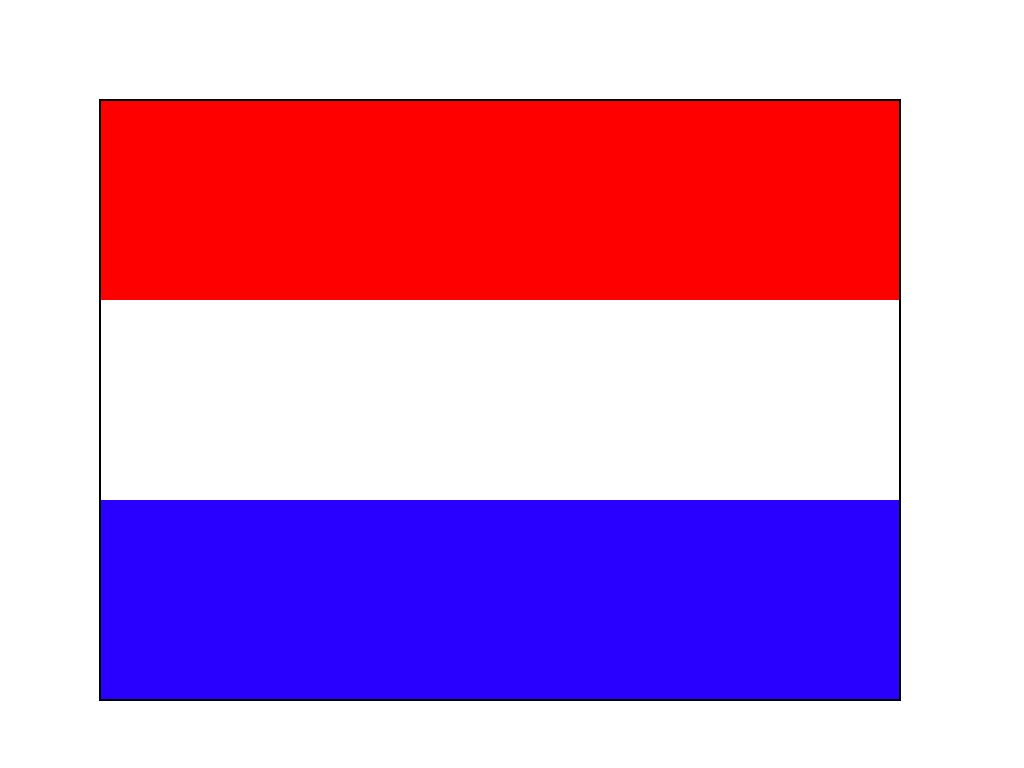} &
  \includegraphics[width=3cm]{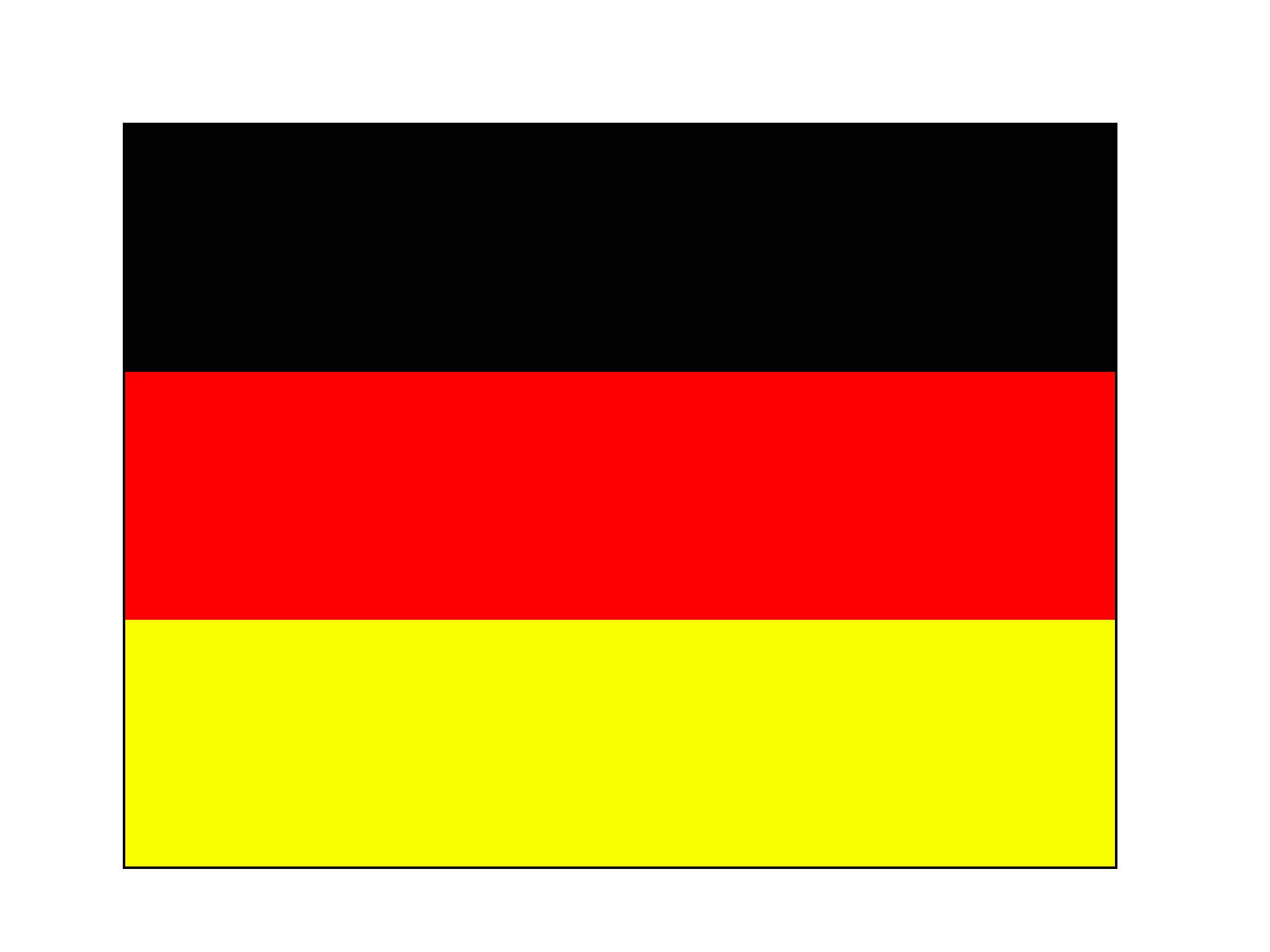} &
   \includegraphics[width=3cm]{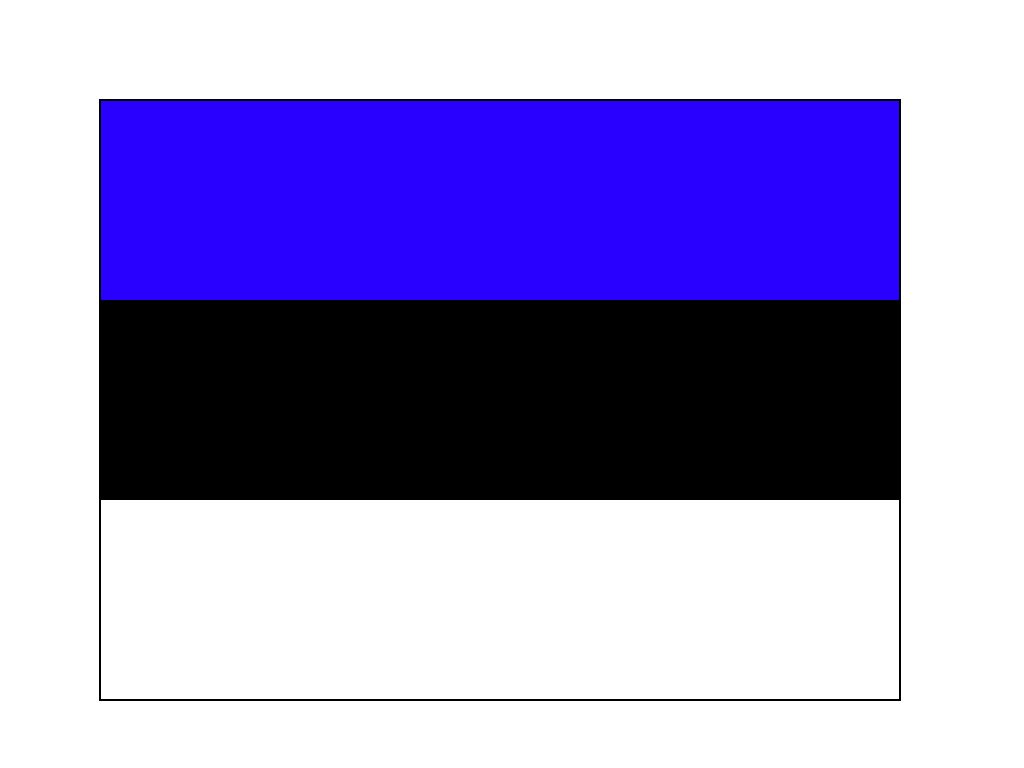} &
  \includegraphics[width=3cm]{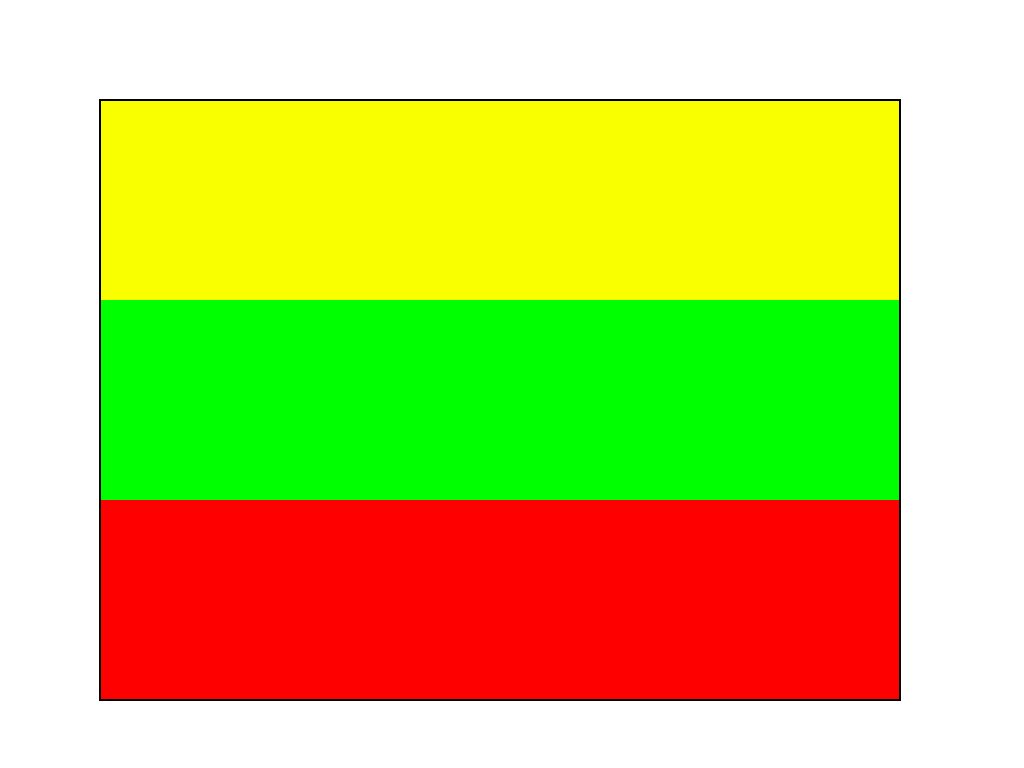} &
   \includegraphics[width=3cm]{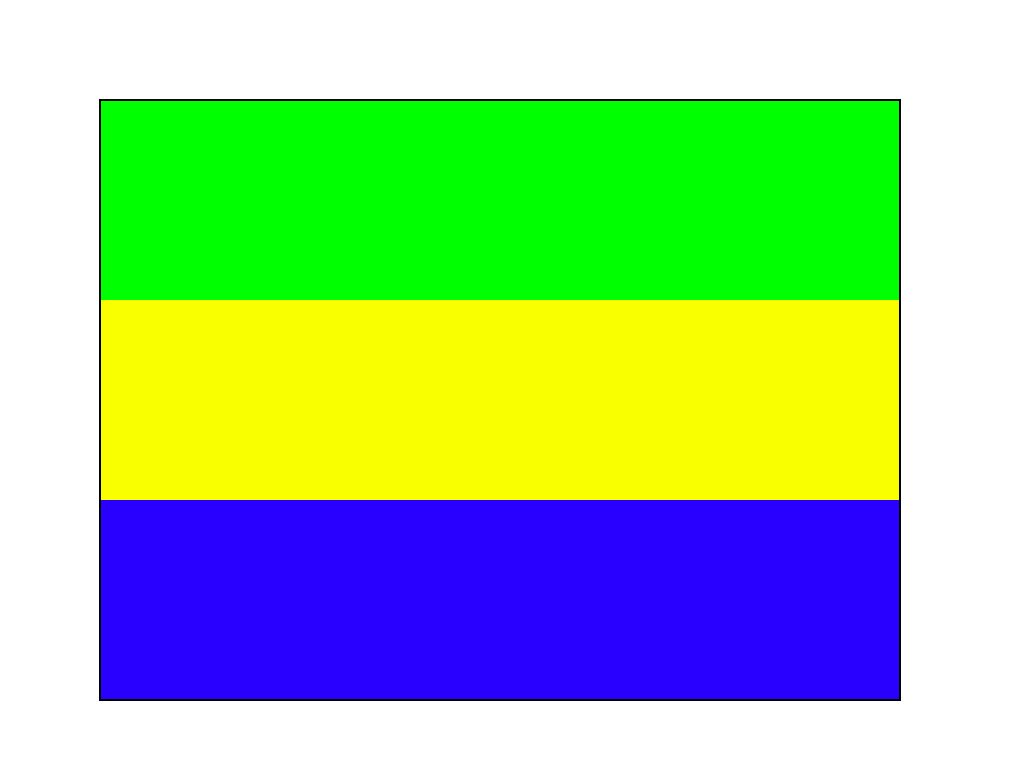}\\
   Netherlands&Germany&Estonia&Lithuania&Gabon\\
  \end{tabular}
    \caption{National Flags \label{fig:flags}}
\end{figure}

%%%%%%%%%%%%%%%
Let us now assume the completely opposite distribution of the coefficients, \ie to represent one element in a class we need to combine all features of that class with equal magnitudes, \ie
\begin{align}
\sig_i= \Fn_i \coef_i + r_i \: \mbox{ with } \: |\coef_i(k)| = x, \mbox{ for }k=1\ldots \nf.
\end{align}
An example would be trying to label pictures of national flags and with the corresponding countries. For simplicity assume that the only flags in question are those of the Netherlands, Germany, Estonia, Lithuania and Gabon, which all consist of three horizontal stripes in various colours, \ie red, white and blue for the Netherlands, black, red and yellow for Germany, blue, black and white for Estonia, yellow, green and red for Lithuania and green, yellow and blue for Gabon, cp. Figure~\ref{fig:flags}. Good features in this example are the colours of the stripes. Each national flag has its three distinctive colours which appear in an equal amount but are not exclusive to this flag.

% 1 column

% 2 columns
%\begin{figure}[htb]
%%\centering
%\begin{tabular}{ccc}
% \includegraphics[width=2.5cm]{nl.jpg} &
%  \includegraphics[width=2.5cm]{de.jpg} &
%   \\
%    Netherlands&Germany& \\
%   \includegraphics[width=2.5cm]{es.jpg} &
%  \includegraphics[width=2.5cm]{lt.jpg} &
%   \includegraphics[width=2.5cm]{gb.jpg}\\
%   Estonia&Lithuania&Gabon\\
%  \end{tabular}
%    \caption{National Flags \label{fig:flags}}
%\end{figure}

In this case we get the maximal response from the correct class when choosing $p=1$, \ie $\|\coef\|_1=sx$. Also from Inequality~\eqref{pq-bound} we see that using $q=\infty$ and $p=1$ gives a very beneficial bound.
\begin{align}
\frac{\|\Fn_j^\star \sig_i\|_1}{\|\Fn^\star_i\sig_i\|_1} &\leq  \frac{\|\Fn_j^\star \Fn_i \|_{\infty,1} }{s} +  \frac{\| \Fn_j^\star r_i\|_1}{sx}.  \notag
\end{align}
Remembering that $\|\Fn_j^\star \Fn_i \|_{\infty,1}$ is smaller than the absolute sum of all the correlations between features in one class and features in another class we get a less sharp version of the above bound,
\begin{align}
\frac{\|\Fn_j^\star \sig_i\|_1}{\|\Fn^\star_i\sig_i\|_1} &\leq  \frac{\sum_{k,l}|\ip{\fn_i^k}{\fn_j^l}| }{s} +  \frac{\| \Fn_j^\star r_i\|_1}{sx}. \notag
\end{align}
which shows that for the case of flat coefficients we have a quite different coherence constraint. Even if a few features in a class are very close to features in another class or actually the same this is not a problem as long as the majority of features from two different classes are not very correlated.\\
In the example of the flags this means that even though two different national flags might share up to two colours, as long as we take into account that all three colours have to appear to the same degree, we can still identify the country from a picture of the flag.

%%%%%%%%%%%%%%%%%%%%%
\subsection{Unstructured Coefficients}
%%%%%%%%%%%%%%%%%%%%%
The last case we are going to discuss is probably the most common and concerns coefficients which follow neither of the two extreme distributions discussed above or where the exact distribution is unknown. An example is the task of face recognition, \ie identifying a person from a picture. Obvious features in this case are noses, eyes and mouths. In any picture most of these features will be visible but their strength will largely depend on the facial expression and lighting conditions. To choose a good p-norm for the classification in this case, we again bound the norm ratio we need to be small.
\begin{align}\label{ppbound}
\frac{\|\Fn_j^\star \sig_i\|_p}{\|\Fn^\star_i\sig_i\|_p} &\leq \frac{\|\Fn_j^\star \Fn_i 
\coef_i\|_p}{\|\coef_i\|_p} +  \frac{\| \Fn_j^\star r_i\|_p}{\| \coef_i\|_p}.
\end{align}
Since we do not have information about the shape of the coefficients, the first term on the right hand side can be as big as $\|\Fn_j^\star \Fn_i \|_{p,p}=\max_{\|x\|_p=1} \|\Fn_j^\star \Fn_i \|_{p}$. Taking into account the orthogonality of the features in the matrices $F_i$, we see that for $p=2$ this term can only be equal to one if two classes overlap, meaning that there is a signal whose features in its own class can be represented by features in a different class. For $p=1/\infty$, however, the corresponding term is equal to the maximum absolute column/row sum of the $\Fn_j^\star \Fn_i$ and it can be easily seen that this can be larger than one, even if for no signal the features in its own class can be fully represented by features in a different class. Similar results hold for all other $p\neq 2$, thus making $p=2$ the best choice in this case. Observe also that $p=2$ corresponds to measuring the energy captured by the features of a class. Thus if the features are well chosen also the second term in Inequality~\eqref{ppbound} can be expected to be small.\\
Finally we see that choosing $p=2$ puts the following incoherence constraint on the feature spaces.  No signal that can be constructed from features in one class should be well representable by features in another class. This constraint is the strongest we have encountered so far, which is only natural since we do not have an assumption on coefficient distribution. Coming back to our example it also corresponds quite naturally to what one would expect from face recognition, ie. that in all pictures enough distinctive features are visible and no matter the lighting condition or facial expression two people can always be uniquely identified from their features.\\

%%%This means that we get the same estimate as in \eqref{eq:inftybound} for non-orthogonal features. The difficulty when giving up the orthogonality constraint is to decide how much correlation between the features makes sense, since at some point assuming a sparse representation in a set of very similar features might be less useful than assuming a nondescript representation in an orthonormal basis of the feature span.\\
Of course there is ample opportunity to develop more class models, assuming different distributions on the coefficients and using more exotic norms. Also one could use different assumptions on the features, \ie non-orthogonal. However, in this paper we will focus on finding a practical way to calculate sensing or feature matrices for classification based on the three main models.

%%%%%%%%%%%%%%%%%%%%%%
\section{Finding Feature/Sensing Matrices \label{sec:find_sensing}}
%%%%%%%%%%%%%%%%%%%%%%
From the analysis in the last section we can derive two types of conditions that the collection of features or subspaces $\Fn_i$ needs to satisfy. The first type describes how features from different classes should interact, \ie the interplay measured in the appropriate matrix norm should be small, and the second type how the features should interact with the training data, \ie the ratio of the response without to within class should be small. The problem with both kinds of conditions is they are not linear and difficult to handle. For instance calculating the $(2,2)$-norm is equivalent to finding the largest singular value and calculating the $(\infty,1)$-norm is even NP-hard. We will therefore start with a very simple approach that will lead to a reasonably fast algorithm, and in the last section point out how to extend it to include more complicated constraints. 
Instead of requiring explicitly that the interplay between features from different classes is small, hereby avoiding to investigate what small means quantitatively, we use the intuition that this should come as free side effect from regulating the interaction with the training data, and simply ask that $\Fn$ is a collection of orthonormal systems $\Fn_i$ each of rank $\nf$. What we would actually like to do about the interaction of the features with the training data is to minimise the ratio between the response of the training data without to within class. However, a constraint involving the ratio is not linear and very hard to handle. We will therefore split it into two constraints that guarantee that the ratio is small if they are fulfilled. The first constraint is that the response within class is equal to a constant $\beta_p$ which we choose to be the maximally achievable value given the rank of the orthonormal systems and $p$. The second constraint is that the response without class is smaller than a constant $\mu_p$, whose dependence on $\nf, p, \ddim$ is more complicated and will be discussed later. Define the two sets $\mathcal{\Fn}_{\nf}$ and $\mathcal{\Fn}_\mu$ as
\begin{align}
\mathcal{\Fn}_{\nf}&:=\{\Fn=(\Fn_1, \ldots,\Fn_c):\: \Fn_i^\star\Fn_i=I_{\nf} \}\notag\\
\mathcal{\Fn}_\mu&:=\{\Fn:\:  \|\Fn^\star_i \sig_i^k\|_p=\beta_p,\,  \notag \\
& \hspace{4em} \|\Fn^\star_j \sig_i^k\|_p \leq \mu_p,\,\forall k,i,j\neq i \},
\end{align}
then our problems could be summarised as finding a matrix in the intersection of the two sets, \ie
$ \Fn \in \mathcal{\Fn}_{\nf} \cap \mathcal{\Fn}_\mu$.
However, since this intersection might be empty, we should rather look for a pair of matrices, each belonging to one set, with minimal distance to each other measured in some matrix norm, eg. the Frobenius norm, denoted by $\|\cdot\|_{\mathbf{2}}$\footnote{We use this notation instead of the more common variant $\|\cdot\|_F$ to avoid confusion.},
\begin{align}
\min \| \Fn_\nf - \Fn_\mu\|_{\mathbf{2}} \mbox{ s.t. } \Fn_\nf \in \mathcal{\Fn}_{\nf},\, \Fn_\mu \in \mathcal{\Fn}_\mu.
\end{align}
One line of attack is to use an alternate projection method, \ie we fix a maximal number of iterations, an initialisation for $\Fn^0_{\nf}$ and then in each iterative step do:
\begin{itemize}
\item find a matrix $\Fn_{\mu}^k \in \argmin_{\Fn\in \mathcal{\Fn}_\mu} \|\Fn^{k-1}_{\nf}-\Fn\|_{\mathbf{2}}$
\item check if $\|\Fn^{k-1}_{\nf}-\Fn_{\mu}^k\|_{\mathbf{2}}$ is smaller than the distance of any previous pair and if yes store $\Fn^{k-1}_{\nf}$
\item find a matrix $\Fn_{\nf}^k \in \argmin_{\Fn\in \mathcal{\Fn}_{\nf}} \|\Fn_{\mu}^k-\Fn\|_{\mathbf{2}}$
\item check if $\|\Fn^{k}_{\nf}-\Fn_{\mu}^k\|_{\mathbf{2}}$ is smaller than the distance of any previous pair and if yes store $\Fn^{k}_{\nf}$
\end{itemize}
If both sets are convex, the outlined algorithm is known as Projection onto Convex Sets (POCS) and guaranteed to converge. Non convexity of possibly both sets, as is the case here, results in much more complex behaviour. Instead of converging, the algorithm just creates a sequence $(\Fn_{\mu}^k,\Fn_s^k)$ with at least one accumulation point. We will not discuss all the possible difficulties here but refer to \cite{trdhhest05}, where all details, proofs and background information can be found and wherein the authors conclude that alternate projection is a valid strategy for solving the posed problem.\\
To keep the flow of the paper, we will not discuss the two minimisation problems that need to be alternatively solved here. The interested reader can find them, including the exact parameter settings in the simulations of the next section, in the appendix. Instead we will discuss how to set the parameters $\beta_p, \mu_p$ and possible choices for the initialisation $\Fn_{\nf}^0$.\\
As mentioned above we choose $\beta_p$ to be the maximally achievable value. An orthonormal system of $\nf$ feature vectors can maximally take out all the energy of a signal,
\begin{align}
\|\Fn_i^\star \sig_i\|_2 \leq\|\sig_i\|_2.
\end{align}
As the signals are assumed to have unit norm, this energy is at most one and we set $\beta_2=1$. The maximal 1-norm of the vector $\Fn_i^\star \sig_i$ of length $\nf$ with energy 1 is $\sqrt{\nf}$. This is attained when all features of one class take out the same energy, \ie the absolute values of the entries in $\Fn_i^\star \sig_i$ are all equal to $1/\sqrt{\nf}$. This leads to $\beta_1=\sqrt{\nf}$. The infinity norm $\Fn_i^\star \sig_i$ corresponds to the maximal inner product between one of the feature vectors and the signal. As both the feature vector and the signals are normalised, this can be at most one and so we set $\beta_\infty=1$.\\
From the discussion in the last section we see that the parameter $\mu$ reflects the incoherence we require between features from different classes. If we have $\ddim \geq \nclass \cdot \nf$, it is theoretically possible to have $\nclass$ subspaces of dimension $\nf$ which are mutually orthogonal to each other, and $\mu$ could be zero. As soon as the above inequality is reversed, because for instance the actual dimension of the span of all features, \ie $rank(\Fn)$, is smaller than $\ddim$, not all subspaces corresponding to the different classes can be orthogonal but will have to overlap. How the size of this overlap, \ie coherence, should be measured, is determined by the choice of $p$-norm for classification. For instance for $p=2$ the coherence was measured by $\|\Fn_j^\star \Fn_i\|_{2,2}$ and from theory about Grassmannian manifolds, see \cite{trdhhest05}, we know that the maximal coherence between two of $\nclass$ subspaces of dimension $\nf$ embedded in the space $\R^\ddim$ can be lower bounded by
\begin{align}\label{eq:grassbound}
\max_{i\neq j} \|\Fn_j^\star \Fn_i\|^2_{2,2} \geq \frac{\nf\cdot \nclass -\ddim}{\ddim(\nclass-1)}.
\end{align}
The problem with setting $\mu$ as above is that we are not controlling the interaction between the sets of features directly but only indirectly over the training data. There the worst case might not be assumed and so $\mu$ as above would be too large. Also for the cases $p=1,\infty$ we do not have a similar bound. Therefore instead of trying to analyse theoretically how to set $\mu$, where we have to deal with too many unknowns, we use the above bound as an indication of order of magnitude and, when testing our scheme on real data, vary the parameter $\mu$.
Lastly for the initialisation for each class we choose the orthogonal system that maximises the energy taken from this class opposed to the energy taken from the other classes, \ie
\begin{align}
\Fn_{s,i}^0 = \argmin_{\Fn_i^\star \Fn_i=I_s} \|\Fn_i^\star \class_i\|^2_{\mathbf{2}} - \sum_{j\neq i}  \|\Fn_i^\star \class_j \|^2_{\mathbf{2}}.
\end{align}
This problem can be easily solved, by considering the rewritten version of the function to minimise,
\begin{align}
\min_{\Fn_i^\star \Fn_i=I_s} \operatorname{trace} \big(\Fn_i^\star ( \class_i\class_i^\star - \sum_{j\neq i }\class_j\class_j^\star    )\Fn_i\big).
\end{align}
If $UDU^\star$ is an eigenvalue decomposition of the symmetric (Hermitian) matrix $\class_i\class_i^\star - \sum_{j\neq i }\class_j\class_j^\star $, then the minimum is attained for $\Fn_{s,i}^0$ consisting of the $\nf$ eigenvectors corresponding to the $\nf$ largest eigenvalues.

%%%%%%%%%%%%%
\section{Testing\label{sec:testing}}
%%%%%%%%%%%%%
To test the proposed scheme we use two face databases, the AR-database, \cite{mabe98} and the extended Yale B database, \cite{yaleb}. First we will test the validity of all three approaches on the AR-database, even though it is intuitively clear that the most appropriate model for faces corresponds to $p=2$. Using the experience from the AR-database we will then run similar tests on the extended Yale B database using only the most appropriate model $p=2$.

\subsection{AR-Database}
For the test we used a subset of images from the AR-database. For each of the 126 people there are 26 frontal images of size $165\times120$ taken in two separate sessions. The images include changes in illumination, facial expression and disguises. For the experiment we selected 50 male and 50 female subjects and for each of them took the 14 images with just variations in illumination and facial expression, neutral, light from the right and left, front light, angry, happy, sleepy. The all together 700 images from the first session were used as training data and the 699 images\footnote{700 minus corrupted image w-027-14.bmp} from the second session for testing. Every image was converted to grayscale and then stored as a 19800 dimensional column vector. The images from the first session were stored in the $19800\times 700$ matrix $\class^1$ and those from the second in the $19800\times699$ matrix $\class^2$. All images (columns) in $\class^1$ were re-scaled to have unit norm. In order to speed up the calculations, we first applied a unitary transform, which does not change the geometry of the problem, but reduces the size of the matrices, \ie we did a reduced $QR$-factorisation decomposing $Y^1$ into the $19800\times700$ matrix $Q$ with orthogonal columns and the $700\times700$ upper triangular matrix $R$ and set $\tilde{Y}^1=Q^\star Y^1=R$ and $\tilde{Y}^2=Q^\star Y^2$. \\
We tested the proposed scheme for all three choices of $p$ and varying values of $\mu_p$ scaling from $0$ to $10\%$ of $\beta_p$ and number of features per class varying from 1 to 7. The choice of the maximal outside-class contribution $\mu_{\max}=0.1\beta_p$ was inspired by the bound in~\eqref{eq:grassbound}. If we take as effective signal dimension $d=700$ and assume that the space should not only accommodate the 100 different people in our training set but all people, \ie we let $\nclass$ go to infinity, the bound approaches $\sqrt{s/d}$ which is $0.1$ if $s=7$ and $0.0378$ if $s=1$. The maximal number of features per class is 7, since we only have 7 test images and so it does not make sense to look for spaces of higher dimension containing all test images. Note also that for $s=1$ the three schemes are the same, so the results are only displayed once. For each set of parameters we calculated the corresponding feature matrix using the algorithm described in the last section on the images from the first session. We then classified the images from the second session using the appropriate $p$-norm. The results are shown in Tables~\ref{tab:l1}, \ref{tab:l2} and~\ref{tab:li}.

\begin{table*}
\centering
\begin{tabular}{|c|ccccccccccc|}\hline
$\nf \backslash \frac{\mu}{\sqrt{s}}$ & 0 & 0.01 & 0.02 & 0.03 & 0.04 & 0.05 & 0.06 & 0.07& 0.08 & 0.09 & 0.1\\ \hline
2  &  60  &  56 &   56  &  57 &   60  &  58 &   60 &   61  &  66 &   64   & 69\\
3  &  52  &  46  &  48 &   46  &  51  &  51 &   53  &  58  &  62 &   61  &  61\\
4  &  62  &  52  &  54 &   55  &  55  &  56 &   56 &   54  &  55 &   57  &  61\\
5  &  64 &   59  &  56  &  56  &  55  &  58 &   61 &   63  &  66 &   68  &  68\\
6  &  61 &   54  &  57 &   54 &   56  &  59  &  62  &  58  &  61 &   71  &  71\\
7  &  57  &  55  &  57  &  55  &  59  &  57 &   58  &  62  &  61 &   68  &  69\\ \hline
\end{tabular}
\caption{Number of misclassified images on the AR-database for $p=1$ and varying values $\nf$ and $\mu$.}
\label{tab:l1}
\end{table*}

\begin{table*}
\centering
\begin{tabular}{|c|ccccccccccc|}\hline
$\nf \backslash \mu$ & 0 & 0.01 & 0.02 & 0.03 & 0.04 & 0.05 & 0.06 & 0.07& 0.08 & 0.09 & 0.1\\ \hline
1  &  57  &  58  &  59   & 58  &  60  &  59 &   59  &  58  &  58  &  58  &  62\\
2  &  51  &  49  &  51  &  51  &  51 &   55  &  57  &  57  &  59  &  58  &  56\\
3  &  47  &  42  &  45  &  50  &  53 &   53  &  54  &  61  &  62  &  61  &  64\\
4  &  46  &  42  &  41  &  41  &  47  &  48  &  51  &  62  &  63  &  61  &  63\\
5 &   48  &  43  &  40  &  44  &  50  &  51 &   52  &  55  &  55  &  59  &  61\\
6  &  49  &  45  &  42  &  45  &  49  &  48  &  51 &   54  &  54  &  57  &  58\\
7 &   45  &  43  &  43  &  43  &  45  &  45  &  48 &   53  &  51  &  54  &  52\\ \hline
\end{tabular}
\caption{Number of misclassified images on the AR-database for $p=2$ and varying values $\nf$ and $\mu$.}
\label{tab:l2}
\end{table*}

\begin{table*}
\centering
\begin{tabular}{|c|ccccccccccc|}\hline
$\nf \backslash \mu$ & 0 & 0.01 & 0.02 & 0.03 & 0.04 & 0.05 & 0.06 & 0.07& 0.08 & 0.09 & 0.1\\ \hline
 2  & 55  &  62 &   59  &  54  &  56  &  52  &  54  &  61  &  63  &  64  &  62\\
 3   &55  &  63  &  58  &  56  &  60  &  58  &  59  &  63  &  65  &  69  &  69\\
 4   &55  &  64  &  60  &  57  &  59  &  58  &  58  &  61  &  67  &  70  &  67\\
 5 &  55  &  60  &  59  &  55  &  58  &  57  &  57  &  60  &  66  &  71  &  69\\
 6  & 55  &  61  &  59  &  54  &  57  &  56  &  56  &  65  &  67  &  72  &  69\\
 7 &  55   & 61  &  59 &   55  &  56  &  54  &  55  &  66  &  66  &  71  &  70\\ \hline
\end{tabular}
\caption{Number of misclassified images on the AR-database for $p=\infty$ and varying values $\nf$ and $\mu$.}
\label{tab:li}
\end{table*}

As we can see we get the best performance for $p=2$, followed by $p=1$ and $p=\infty$. This comes as no surprise when considering the structure of our data. Intuitively the important features of a face are eyes, nose and mouth. Since the people in the pictures have different facial expression, usually not all of these features will be active explaining why $p=1$ is not the most appropriate model. On the other hand 
we can expect to have more than one feature active at the same time even if not to the same extent. Using $p=\infty$ we lose the information given by these secondary active features while with $p=2$ we still incorporate it into the final decision.\\
We can also see that $0.1\%$ of $\mu$ as maximally allowed outside class 'energy' seemed to have been a good choice as we can always see a small decrease and large increase of the error going from $0$ to $0.1$, with the best range for $p=1$ and $p=2$ between $0.01$ and $0.03$ and for $p=\infty$ between $0.02$ and $0.06$. For $p=1$ we get better performance for the lower dimensions, which seems reasonable because there the equal energy distribution over the features is easier achieved. For $p=2$ on the other hand the better performance is achieved with higher dimensions, which are able to capture more important side details. Finally for $p=2$ the results seem equal for all dimensions. A possible explanation is given by the initialisation, which ensures that for all dimensions the first, most promising direction is included.\\
Still in all three cases in the most promising ranges the proposed scheme outperforms a standard method like Fisher's LDA, \cite{lda}. The best result by LDA is obtained when using the original (not-normalised) images and the highest possible number of discriminant axes $\nclass-1=99$. In this case nearest neighbour classification, corresponding to $p=\infty$ but with non orthogonal features, fails to identify 59 images, and nearest subspace classification, corresponding to $p=2$ fails to identify $71$ images.  
When concentrating on the results for $p=2$, which is the most sensible choice given the structure of the data, $p=2$, we also see that the scheme performs well in comparison to a recent, successful method based on $\ell_1$ minimisation, \cite{wrma09}. The best result reported there is a success rate of $94.99\%$, meaning $35$ misclassified images, which is 5 images better than our best case of $40$ errors.\\
Encouraged by the promising results we now turn to testing our scheme on the extended Yale B database.

%%%%%%%%%%%%%%%%%%
\subsection{Extended Yale B Database}
%%%%%%%%%%%%%%%%%%
From the extended Yale B database we used the 2414 frontal face images, about 64 images taken under varying illumination conditions for each of the 38 people. For the test we randomly split the set of images per person into an equal number of training and test images, using one more training than test image in case of an odd number of images per class. We then ran our classification scheme with the number of features per class varying from 2 to 5 and thanks to the experience gained from the AR-database with the values of $\mu$ running only from 0 to 0.05. For the computation of the feature matrices we used the same simplifications as described for the AR-database. For comparison we ran Fisher's LDA with 37 and 30 discriminative axes in combination with the nearest neighbour classifier. This procedure was repeated 19 times and the mean of all 20 runs was computed. \\ 

\begin{table*}
\centering
\begin{tabular}{|c|cccccc|}\hline
$\nf \backslash \mu$ & 0 & 0.01 & 0.02 & 0.03 & 0.04 & 0.05\\ \hline
2  &  19.80 $\pm$ 5.74 &   20.30 $\pm$ 5.80 &   22.25 $\pm$ 7.20 &   23.85 $\pm$ 6.81 &   25.25 $\pm$ 6.66 &   26.25 $\pm$ 6.61\\
3  &  14.15 $\pm$ 4.37 &   13.60 $\pm$ 4.22 &   13.85 $\pm$ 3.73 &   15.85 $\pm$ 5.25 &   16.40 $\pm$ 4.78 &   17.55 $\pm$ 6.00 \\
4  &  15.75 $\pm$ 3.82 &   14.05 $\pm$ 3.49 &   13.95 $\pm$ 3.55 &   15.35 $\pm$ 3.95 &   16.45 $\pm$ 4.10 &   16.95 $\pm$ 4.30 \\
5 &   15.70 $\pm$ 4.78 &   15.00 $\pm$ 4.91 &   14.45 $\pm$ 4.30 &   15.30 $\pm$ 3.34 &   17.60 $\pm$ 4.65 &   17.65 $\pm$ 4.55 \\ \hline
\end{tabular}
\caption{Mean $\pm$ standard deviation of misclassified images on the Extended Yale B database for $p=2$ and varying values $\nf$ and $\mu$.}
\label{tab:yalebl2}
\end{table*}

The results of our method can be found in Table~\ref{tab:yalebl2}. While Fisher's LDA on average missclassified 23.30 $\pm$ 6.42 images (success rate of 98.07 $\pm$ 0.53\%) using 37 discriminant axes and 231.55 $\pm$ 23.48 images (success rate 80.78 $\pm$ 1.95\%) using 30 discriminant axes, our method in the best case only misclassified 13.60 $\pm$ 4.22 images (success rate 98.87 $\pm$ 0.35\%). In general it outperformed Fisher's LDA for a wide range of values for $\mu$ and $s$.\\
Comparison to the $\ell_1$-minimisation scheme in \cite{wrma09} is harder, as it seems that there only a single run was used. However, their best success rate of 98.26\%, achieved at the same time as Fisher's  LDA with 30 discriminant axes achieved 87.57\% (the maximal rate for Fisher's LDA we encountered in 20 runs was 84.73\%), is still below our best average rate of 98.87\%.\\

\begin{figure}[htb]
\centering
\begin{tabular}{ccc}
 \includegraphics[width=5cm]{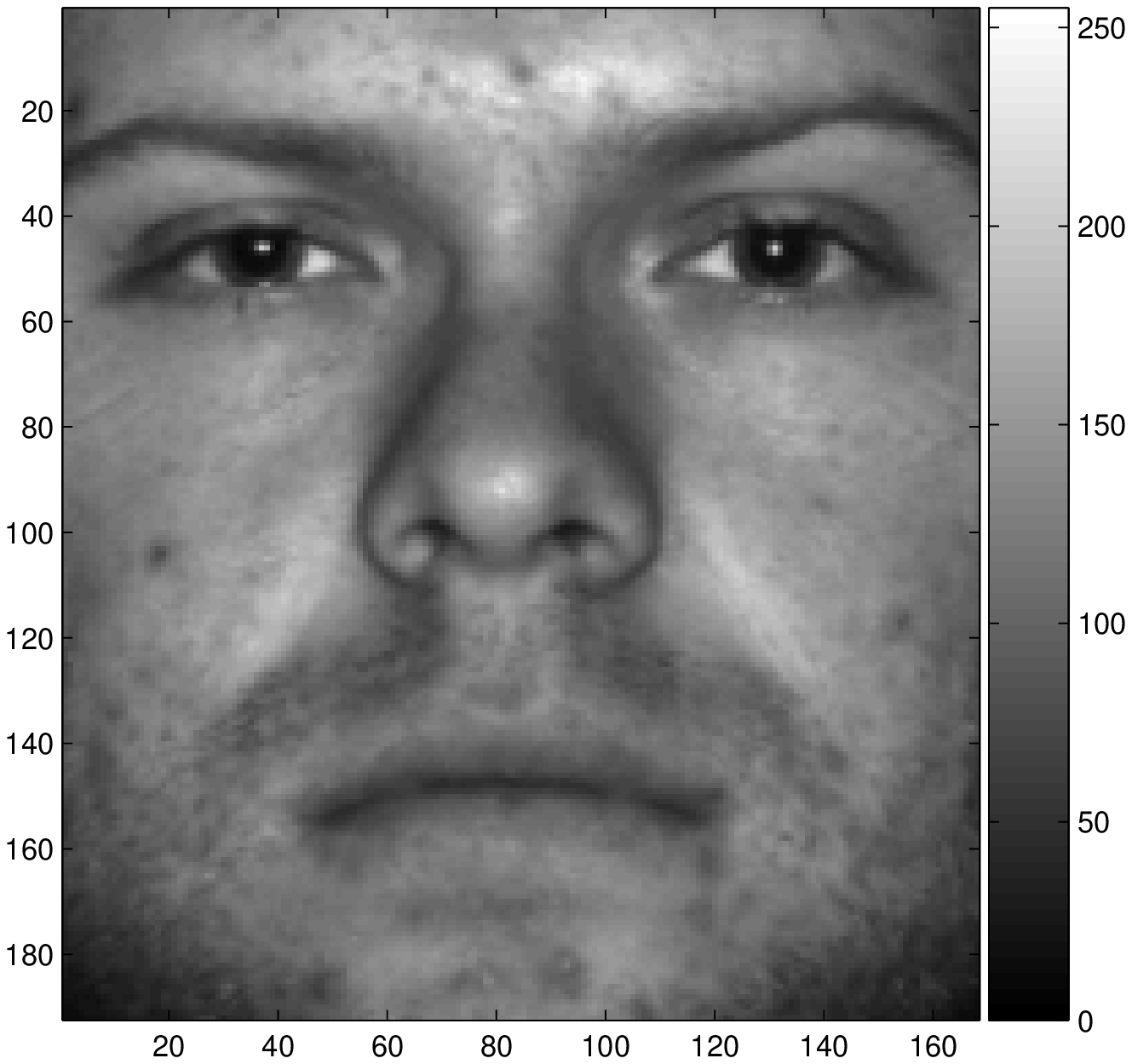} &
  \includegraphics[width=5cm]{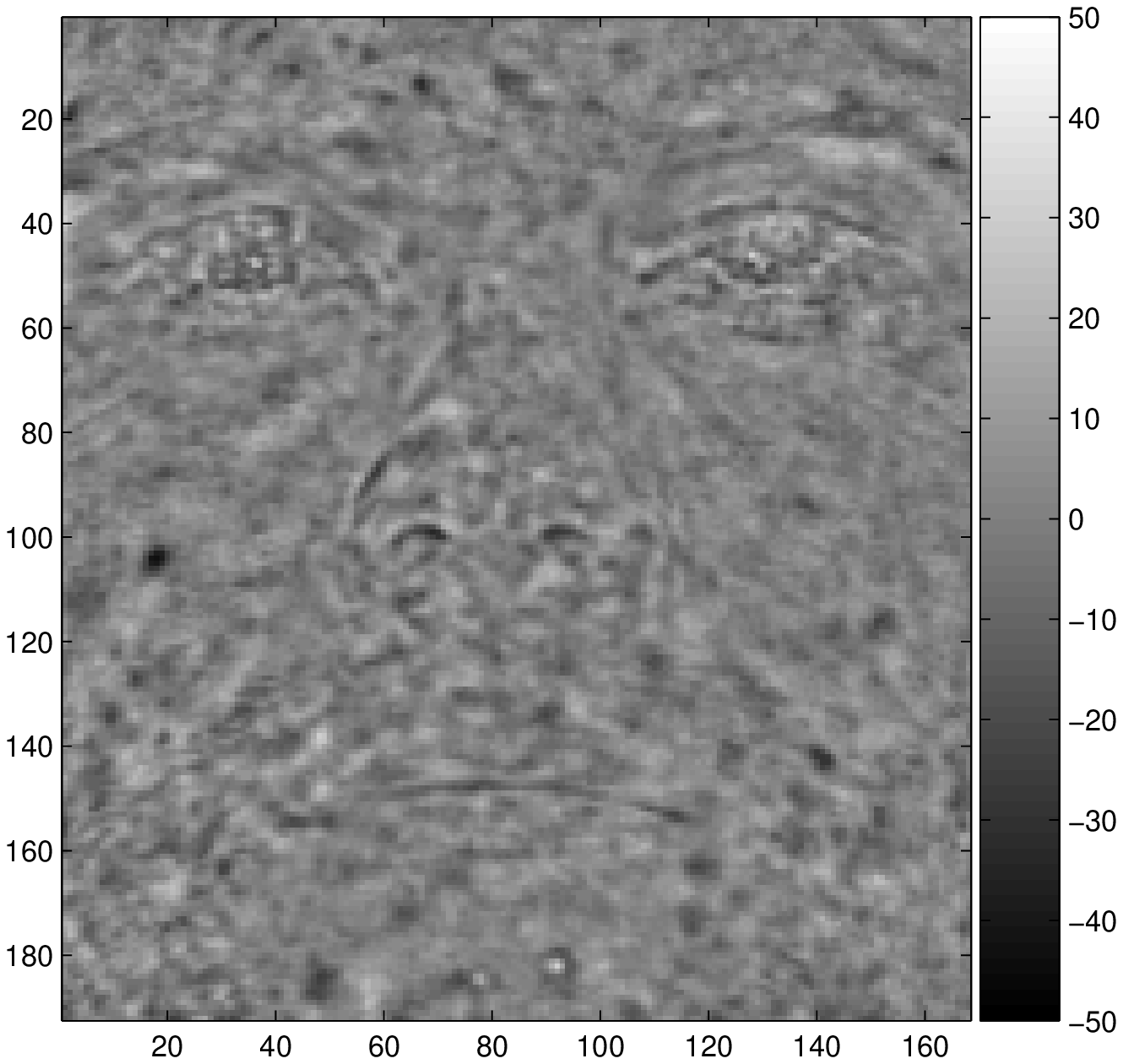} &
   \includegraphics[width=5cm]{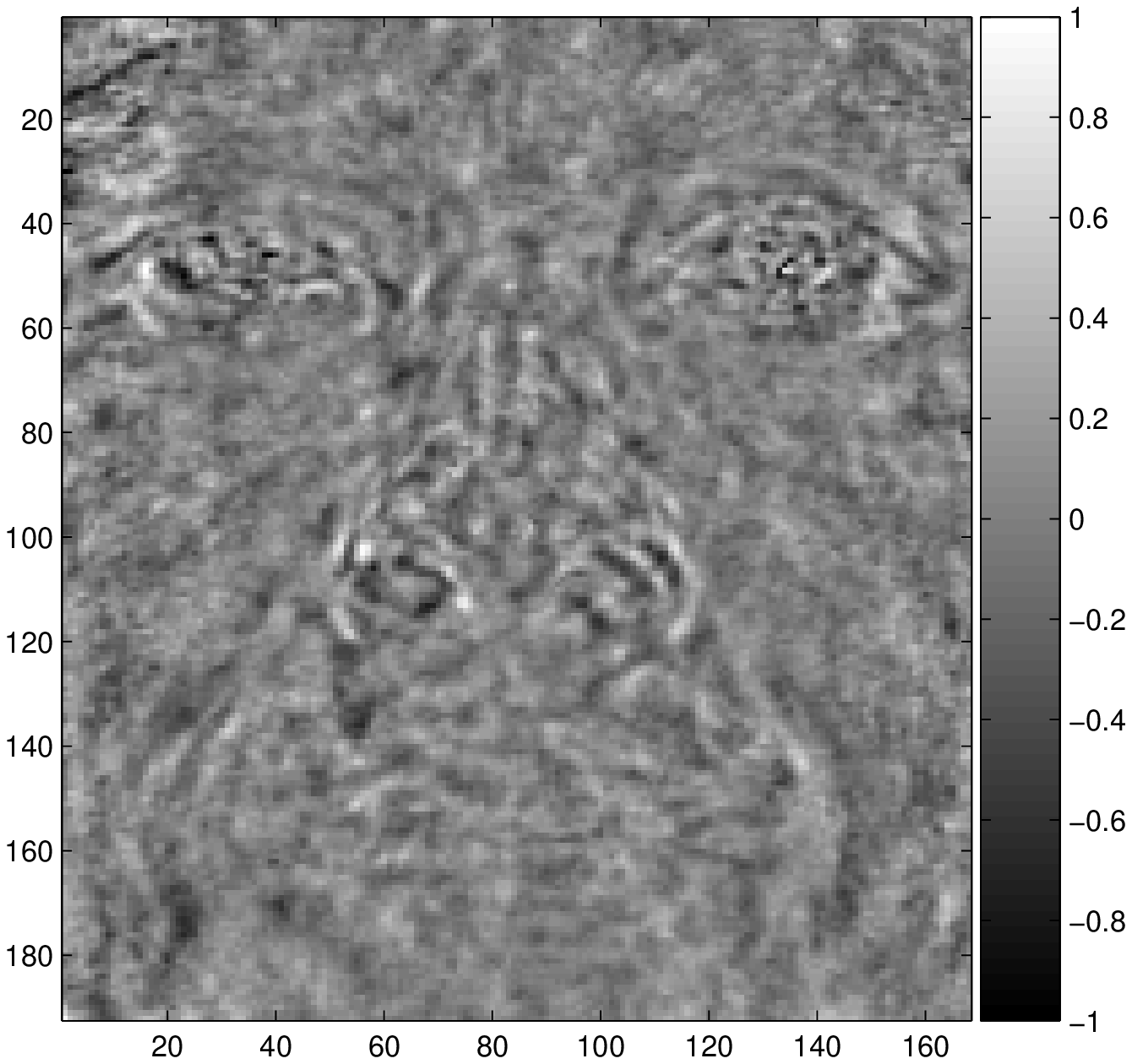}\\
   (a)&(b)&(c)\\
   %$\|y_1\|_2=2.1543\cdot 10^4$ &$\|\Fn_1 \Fn_1^\star y_1\|_2=1.2073\cdot 10^3$ &$\|\Fn_2 \Fn_2^\star y_1\|_2=32.6198$\\
% Gabor Dictionary\\ 
 \includegraphics[width=5cm]{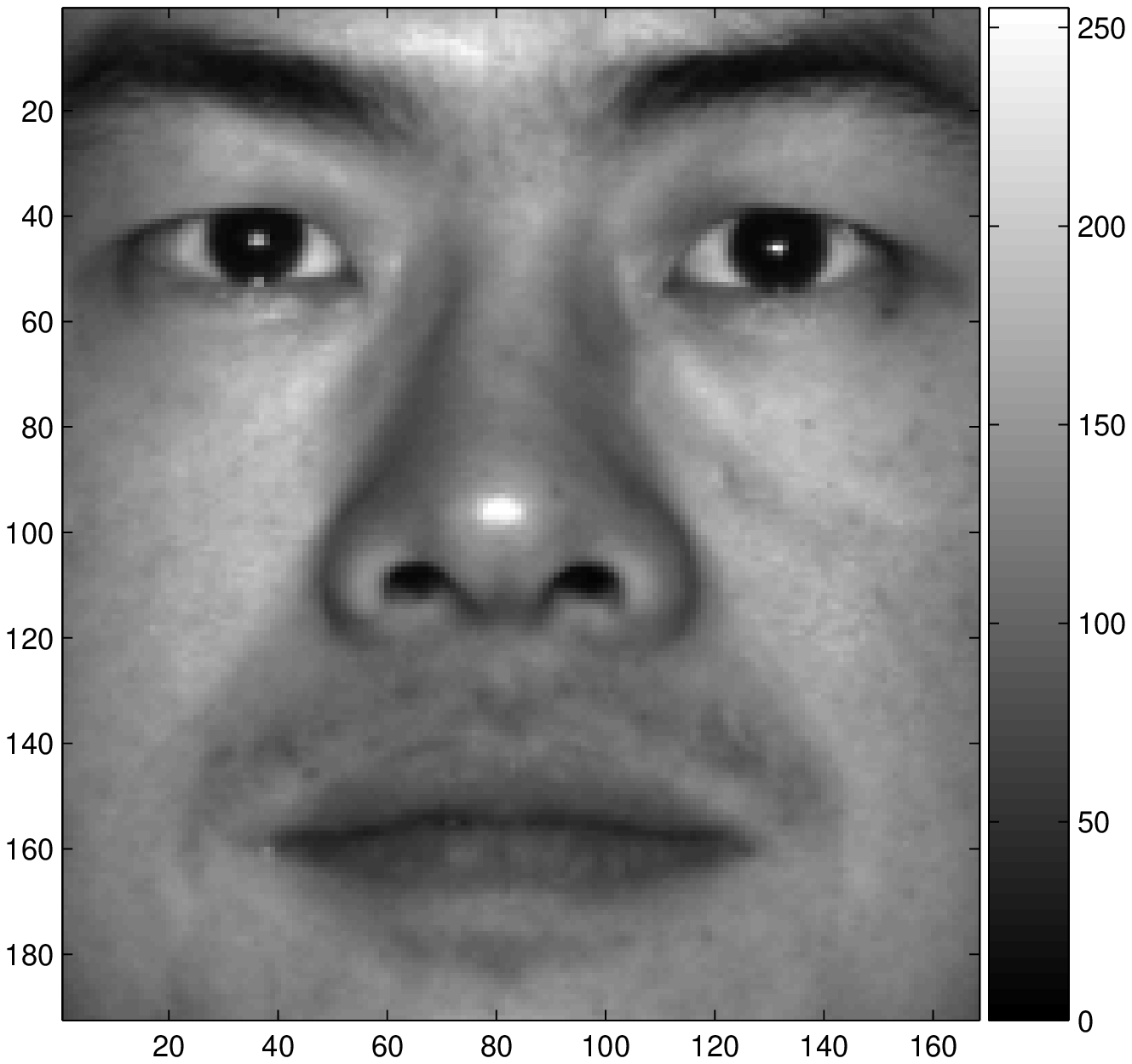} &
  \includegraphics[width=5cm]{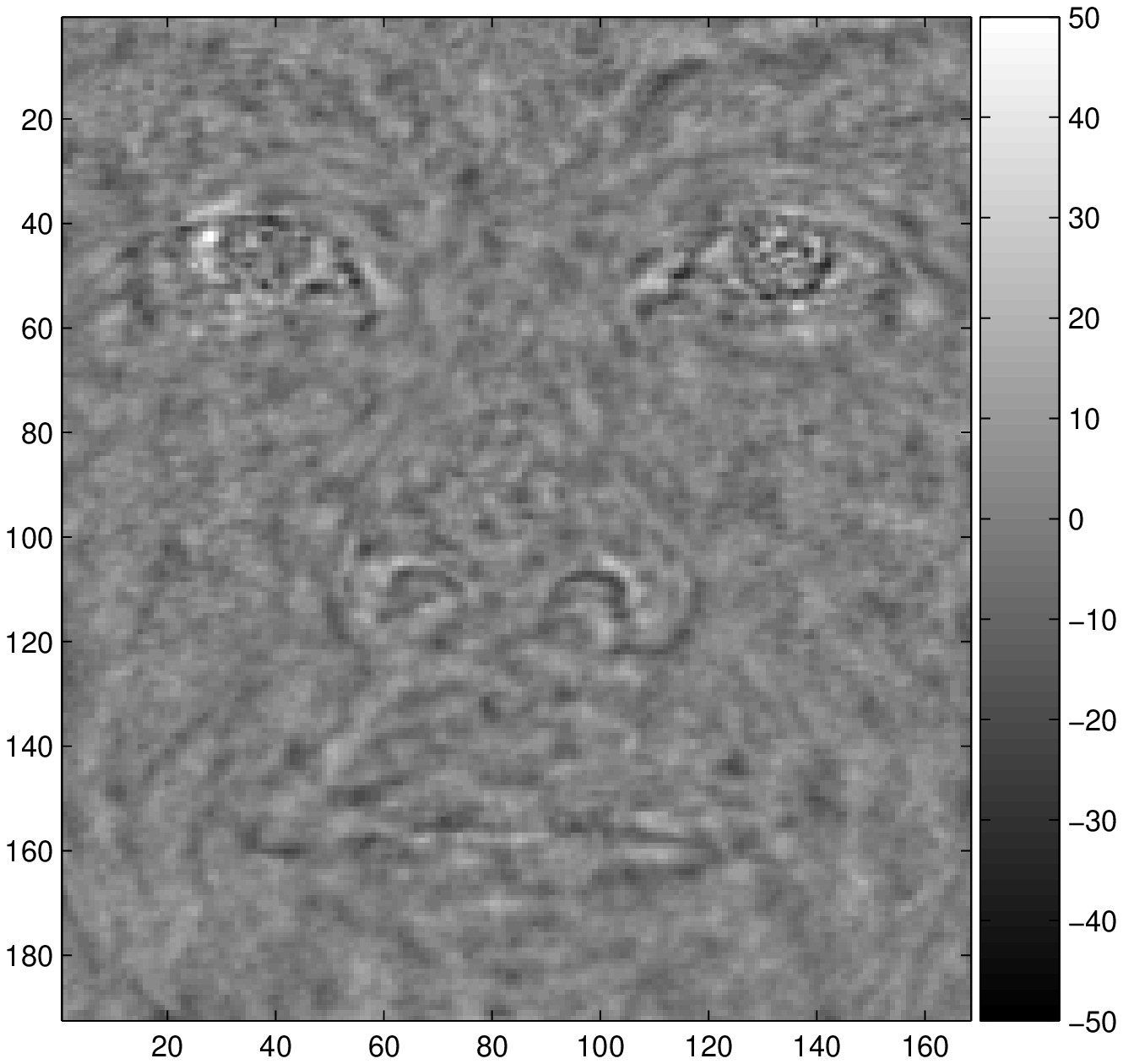} &
   \includegraphics[width=5cm]{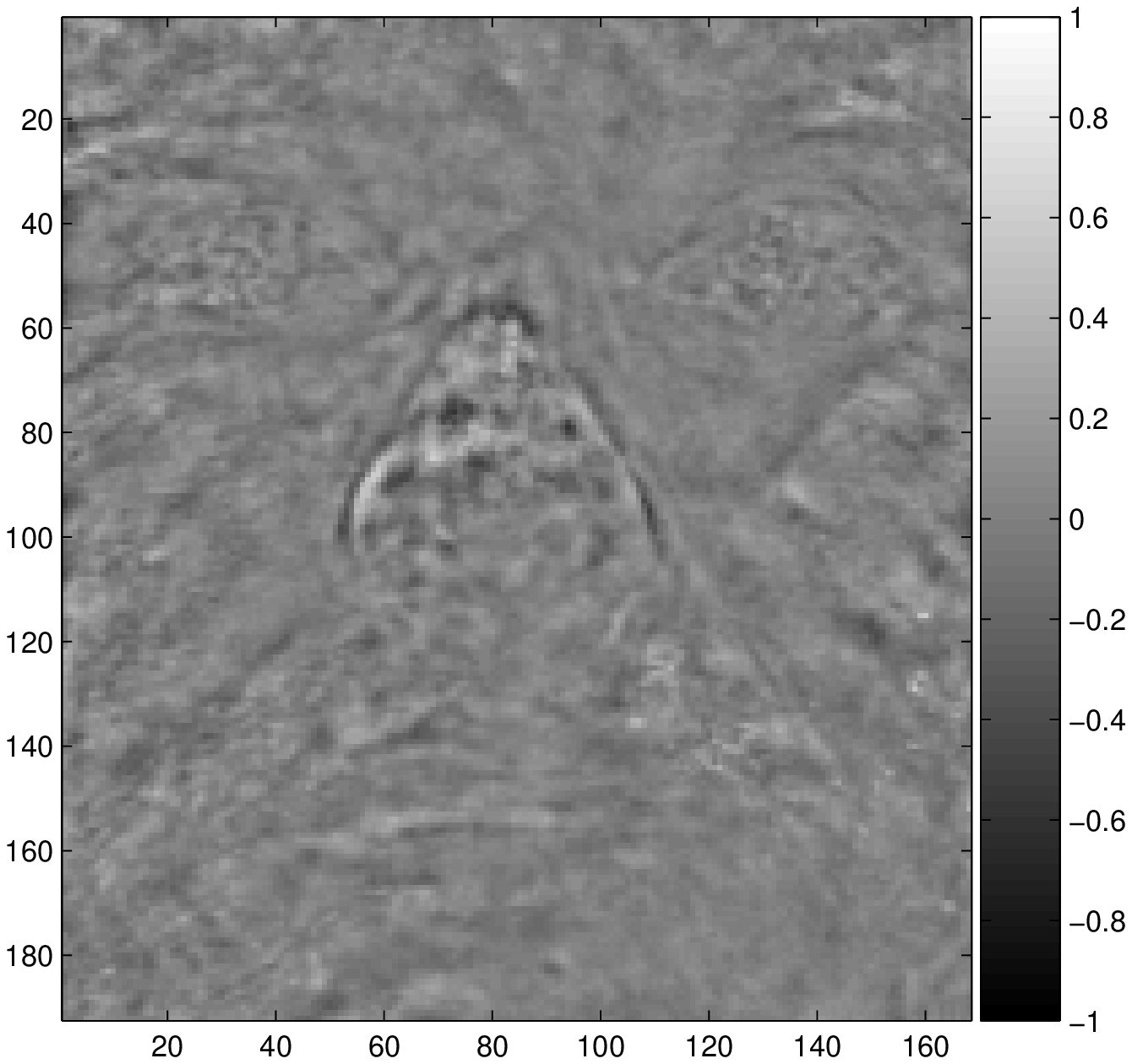}\\
  (d)&(e)&(f)
      %$\|y_2\|_2=2.2472\cdot 10^4$ &$\|\Fn_2 \Fn_2^\star y_2\|_2=1.0808\cdot 10^3$ &$\|\Fn_1 \Fn_1^\star y_2\|_2=17.5149$\\
  \end{tabular}
    \caption{Images of two subjects, original (a) \& (d), projected onto the span of features from their own class (b) \& (e), projected onto the span of features of the wrong class (c) \& (f) \label{fig:faces}}
\end{figure}

To illustrate the results in Figure~\ref{fig:faces} and to confirm the motivation in the introduction for using different features for different classes, we show what happens to the training images of two different subjects when projected on the features of their own class and the other subject's class. As expected the projections on features of their own class nicely filter out common traits like eyes, mouths and noses, but on top of that the features of the first subject capture the very distinctive birth mark on his right cheek. The projections on the wrong class on the other hand are not only much weaker (note the difference in scale) but also less clear. Two overlapping sets of features seems to appear at the same time, the ones that belong to the subject in the image and the ones that the projection is trying to filter out. 

% 1 column 
 
% 2 columns
%
%\begin{figure}[htb]
%\centering
%\begin{tabular}{ccc}
% \includegraphics[width=2.5cm]{sub1.eps} &
%  \includegraphics[width=2.5cm]{sub11.eps} &
%   \includegraphics[width=2.5cm]{sub12.eps}\\
%   (a)&(b)&(c)\\
%   %$\|y_1\|_2=2.1543\cdot 10^4$ &$\|\Fn_1 \Fn_1^\star y_1\|_2=1.2073\cdot 10^3$ &$\|\Fn_2 \Fn_2^\star y_1\|_2=32.6198$\\
%% Gabor Dictionary\\ 
% \includegraphics[width=2.5cm]{sub2.eps} &
%  \includegraphics[width=2.5cm]{sub22.eps} &
%   \includegraphics[width=2.5cm]{sub21.eps}\\
%  (d)&(e)&(f)
%      %$\|y_2\|_2=2.2472\cdot 10^4$ &$\|\Fn_2 \Fn_2^\star y_2\|_2=1.0808\cdot 10^3$ &$\|\Fn_1 \Fn_1^\star y_2\|_2=17.5149$\\
%  \end{tabular}
%    \caption{Images of two subjects, original (a) \& (d), projected onto the span of features from their own class (b) \& (e), projected onto the span of features of the wrong class (c) \& (f) \label{fig:faces}}
%\end{figure}

%(3dim per class 0.01)
% norm(im1,'fro')=   2.1543\cdot 10^4
%norm(im2,'fro')= 2.2472\cdot 10^4
%norm(im11,'fro')=1.2073\cdot 10^3
%norm(im22,'fro')=1.0808\cdot 10^3
%norm(im12,'fro')=32.6198
%norm(im21,'fro')=17.5149

Summarising the results, we can say that our method outperforms a classic scheme like Fisher's LDA. In comparison to the $\ell_1$-minimisation scheme in \cite{wrma09} it performs slightly worse on the AR-database but seem to be better on the YaleB-database. However it has one big advantage over the $\ell_1$-minimisation scheme, which is its low computational complexity. Not taking the calculation of the feature matrices into account, as this is part of the pre-processing, basically all that has to be done to classify a new data vector is to multiply it with the feature matrix and calculate some statistics on the resulting vector. The $\ell_1$ minimisation method on the other hand requires on top of extracting the features the solution of a convex optimisation problem
\begin{align}
\min \|z\|_1  \mbox{ s.t. } \| f_{new} - F z\|_2\leq \eps,
\end{align}
where $F$ in this case is the $d_f \times \nsig$ matrix containing the features of all the training data. 
For comparison in \cite{wrma09} the authors state that the classification of one image takes a few seconds on a typical 3 GHz Pc.  At the same time for classifying 1205 images of size $192\times 168$, using our method with 4 feature dimensions per class, MATLAB takes less than half a minute on a Dual 1.8Ghz PowerPC G5, which is less than 25ms per image.

In the next section we will introduce the mathematical framework on which we base our classification scheme. It consists of a model of subspaces, one associated to each class, and a model on how the elements in this class are represented in this subspace, which together lead to a natural choice of the norm we have to use for the classification and an incoherence requirement on the subspaces. 
%%%%%%%%%%%%%%%%%%%%%%%%%
\section{Discussion}
%%%%%%%%%%%%%%%%%%%%%%%%%
We have presented a classification scheme based on a model of incoherent subspaces, each one associated to one class, and a model on how the elements in a class are represented in this subspace.  From a more practical viewpoint we have developed an algorithm to calculate these subspaces, \ie the feature matrices, and shown that the scheme gives promising results on the AR database and even outperforms a state of the art method like the $\ell_1$-minimisation scheme in \cite{wrma09} on the YaleB-database. The idea that each class should have its own representative system, learned from the training data can already be found in \cite{skhu06}. There frames or dictionaries for texture classification are learned, such that each provides a sparse representation for its texture class. The new texture then gets the label of the texture frame providing the sparsest representation. In \cite{masazi08}, the same basic idea is used but the learning is guided by the principle that the dictionaries should also be discriminant, while in \cite{rosa08} both learning principles are combined, \ie the dictionaries should be discriminant and approximative. This third scheme can be considered as a more general and more complicated version of our approach. Alternatively our approach can be considered to be a hybrid of Nearest Subspace respectively Nearest Neighbour and the discriminative and approximative frame scheme, in so far as it is linear but has individual features for every class. \\
The idea to use a collection of subspaces for data analysis can also be found in \cite{Mayadefo08}, where the subspaces are used to model homogenous subsets of high-dimensional data which together can capture the heterogenous structures.\\
For the future there remain some interesting directions to explore. Firstly the possibilities of the subspace classification approach do not seem exhausted using the proposed algorithm. Ironically this fact revealed itself through a bug in the minimisation procedure, resulting in matrix pairs with distances larger than the optimal ones, and sensing matrices giving better classification results, \ie in the best case an error of only $35$ misclassified images. The main difference of these fake optimal matrices to the sensing matrices corresponding to the actual minima, seemed to be that, while capturing approximately the same 'energy' within class, they were more accurate in respecting the without class energy bound, \ie less overshooting of the maximally allowed value $\mu$. This overshooting for the real minima is a result of imposing not only $\|\Feat_i \sig_j^k\|_2\leq\mu$ but also $\|\Feat_i \sig_i^k\|_2=\beta$, which forces the optimal feature matrix to balance the error incurred by not attaining $\beta$ within class and the error incurred by being larger than $\mu$ without class.
A promising idea to avoid the overshooting would be to change the problem formulation and ask to
maximise the 'energy' within class subject to keeping the 'energy' without class small, \ie in the case $p=2$ solve,
\begin{align}
\max \sum_i \|\Feat_i^\star \class_i&\|^2_\mathbf{2} \notag \\
 \mbox{ s.t. }  & \Feat_i^\star \Feat_i = I_\nf  \mbox{ and } \notag \\
&\|\Feat_i x_j^k\|_2 \leq \mu, \,\forall k, j\neq i.\notag
\end{align}
Lastly our approach allows to impose additional constraints on $\Feat$, like incoherence of the subspaces between each other, e.g. $ \|\Feat_i^\star \Feat_j\|_{2,2}\leq \nu$ for $p=2$, or low rank of the whole feature matrix to reduce the cost of calculating $\Feat^\star \sig_{new}$. Another possibility to reduce computational cost if $\ddim$ and $\nsig$ are very large, especially in the training step, would be to first take random samples of the training data, which reduce their dimension but very likely preserve the geometrical structure, as described in \cite{ac01} and used in \cite{wrma09}. Alternatively to reduce the dimension of $F$ one can apply our scheme on classical features, like Eigen or Laplace features, instead of directly on the raw training data.

%%%%%%%%
\appendix
%%%%%%%%
\section{Solution Sketches for the Minimisation Problems}
In order to use the alternate projection method for calculating the feature matrices we need to find the projection of a matrix $\hat{\Feat}$ onto $\mathcal{\Feat}_s$ and onto $\mathcal{\Feat}_\mu$ in the three cases $p=1,2,\infty$.
We will start with the easier of the two problems
\begin{align}
\mbox{find: } \Feat_s\in \argmin_{\Feat \in \mathcal{\Feat}_s } \|\Feat-\hat{\Feat}\|_{\mathbf{2}}.
\end{align}
Since the minimisation problem is invariant under squaring of the objective function and thus equivalent to
\begin{align}
\min_{\Feat \in \mathcal{\Feat}_s } \|\Feat-\hat{\Feat}\|^2_{\mathbf{2}}=
\min_{\Feat \in \mathcal{\Feat}_s } \sum_{i=1}^\nclass \|\Feat_i-\hat{\Feat}_i\|^2_{\mathbf{2}},
\end{align}
it splits into $\nclass$ independent problems
\begin{align}
\min_{\Feat_i^\star\Feat_i = I_{\nf} } \|\Feat_i-\hat{\Feat}_i\|^2_{\mathbf{2}}.
\end{align}
The solution of these problems is straightforward. If $\hat{\Feat}_i$ has the reduced singular value decomposition $\hat{\Feat}_i= U_iS_iV_i$ then the orthonormal system $\Feat_i$ of same rank closest to it is $\Feat_i=U_iV_i$, see e.g. \cite{horn_johnson}.\\
The second minimisation problem 
\begin{align}
\mbox{find: } \Feat_\mu\in \argmin_{\Feat \in \mathcal{\Feat}_\mu } \|\Feat-\hat{\Feat}\|_{\mathbf{2}}. \notag
\end{align}
is more complicated to solve. Assume that the number of training signals is larger than the dimension of the signals and span the whole space, so that the $\ddim\times \nsig$ matrix $Y$ has rank $\ddim \leq \nsig$. If not we embed the training signals into a lower dimensional space corresponding to the rank of $\class$ via a reduced $QR$-decomposition of $\class$ and set $\tilde{\class}=Q^\star \class=R$ before starting the alternating projection procedure. Afterwards we set $\Feat=Q^\star \tilde{\Feat}$, where $\tilde{\Feat}$ is the feature matrix calculated from the lower dimensional embedded data.
Since $\class$ has rank $\ddim$ we have $\class \class^\dagger = I_d$ and can reformulate the problem to solve as
\begin{align}\label{eq:strucF}
\min_{\Feat \in \mathcal{\Feat}_\mu } \|\Feat-\hat{\Feat}\|_{\mathbf{2}} = \min_{\Feat \in \mathcal{\Feat}_\mu } \|(\Feat^\star \class - \hat{\Feat}^\star\class)\class^\dagger\|_{\mathbf{2}}.
\end{align}
The advantage of this formulation is that it is in terms of $\Feat^\star \class$, which is also used to describe $\mathcal{\Feat}_\mu$. To further exploit this property we define the set $\mathcal{G}_\mu$, which is of the form $\Feat^\star \class$ with $\Feat \in \mathcal{F}_\mu$. To characterise the set $\mathcal{G}_\mu$ we assume the following notation. Let $G_{ij}$ refer to the $\nf \times n_j$ submatrix that corresponds to $\Feat_i^\star \X_j$ inside $\Feat^\star \X$ and denote the $k$-th column of $G_{ij}$ by $G_{ij}(:,k)$. We can then define
\begin{align}\label{eq:defGmu}
\mathcal{G}_\mu:=\{G: \:&\|G_{ii}(:,k)\|_p=\beta_p, \, \notag\\
&\|G_{ij}(:,k)\|_p\leq \mu_p, \forall k,i,j\neq i \}. 
\end{align}
Set $\hat{G}=\hat{\Feat}^\star \class$ then the problem in \eqref{eq:strucF} is equivalent to
\begin{align}\label{eq:strucG}
\min_{G \in \mathcal{G}_\mu } \|(G - \hat{G})\class^\dagger\|_{\mathbf{2}}.
\end{align}
To attack this problem we will use resolvents or proximity operators which are a generalisation of projection operators. Given a Hilbertspace $\mathcal{H}$  and a function $f$ from $\mathcal{H}$ to $]-\infty,+\infty]$ that is lower semicontinuous, convex and not identical to $+\infty$, \ie belonging to $\Gamma_0(\mathcal{H})$ the proximity operator $\prox_f$ is defined by
\begin{align}
\prox_f(x)= \argmin_{\mathcal{H}} f(y) + \frac{1}{2} \|x-y\|^2_\mathcal{H}. \notag
\end{align}
Proximity operators were first studied by Moreau in \cite{mo62}, who developed a theory of proximal calculus, and recently have been used to solve optimisation problems in signal processing, \cite{cope07}. Here we will use the forward backward splitting approach as described in \cite{cowa05}. Assume that we can write the function to minimise as the sum of two functions $f_1, \, f_2$ in $\Gamma_0(\mathcal{H})$, \ie 
\begin{align}\label{eq:proxprob}
\min_{x \in \mathcal{H}} f_1(x) + f_2(x).
\end{align}
If $f_2$ is differentiable with a $\beta$-Lipschitz continuous gradient for $\beta
>0$ then the sequence generated by fixing $x_0 \in \mathcal{H}$ and iterating
\begin{align}\label{eq:iteration}
x^{n+1}=\prox_{\gamma^n f_1} (x^{n} -\gamma^n \nabla f_2(x^n))
\end{align}
converges weakly to a minimum of~\eqref{eq:proxprob} if $\gamma < 2/\beta$.\\
To apply the concept to our problem we take as Hilbert space the set of all $\nclass \cdot \nf \times \nsig$ matrices $G$ equipped with the Frobenius norm and define the indicator function $\mathbb{I}_{\mathcal{G}_\mu}$ of the set $\mathcal{G}_\mu$ by
\begin{equation}
\mathbb{I}_{\mathcal{G}_\mu} (G) :=  \left\{ \begin{array}{cl} % brackets may be (...), [...], \{...\}, or left out
      1 & \mbox{ if } G\in \mathcal{G}_\mu\\
     + \infty & \mbox{ else} \\
   \end{array} \right. . \notag
\end{equation}
The we can replace problem \eqref{eq:strucG} by
\begin{align}\label{eq:strucI}
\min_{G } \mathbb{I}_{\mathcal{G}_\mu}(G) + \|(G - \hat{G})\class^\dagger\|^2_{\mathbf{2}} .
\end{align}
The slight imperfection of this approach is that the set $\mathcal{G}_\mu$ is not convex, therefore $ \mathbb{I}_{\mathcal{G}_\mu}(G)$ is not convex and the sequence generated applying~\eqref{eq:iteration} is not guaranteed to converge. Finding only a local minimum is however not such a big problem, since the procedure is only part of a bigger iterative scheme, as long as in each step we get some improvement.\\
What remains to be done is to calculate the proximity operators for $\gamma f_1=\gamma \mathbb{I}_{\mathcal{G}_\mu} = \mathbb{I}_{\mathcal{G}_\mu} $, the gradient of $f_2(G)=\|(G - \hat{G})\class^\dagger\|^2_{\mathbf{2}}$ and decide about the initialisation $G_0$ and the step sizes $\gamma^n$. A straightforward calculation shows that $\nabla f_2(G)= 2(G-\hat{G}) Y^\dagger (Y^\dagger)^\star$.
Since $ \mathbb{I}_{\mathcal{G}_\mu} $ is an indicator function the proximity operator is simply the orthogonal projection onto $\mathcal{G}_\mu$, \ie
\begin{align}
\argmin_{G} \mathbb{I}_{\mathcal{G}_\mu}(G)  + \frac{1}{2} \|G^n-G\|^2_\mathbf{2} = \argmin_{G\in \mathcal{G}_\mu} \|G^n-G\|^2_\mathbf{2} \notag
\end{align}
Because of the structure of $\mathcal{G}_\mu$, see \eqref{eq:defGmu}, the problem above splits into the smaller problems
\begin{align}
&\min_{\|G_{ii}(:,k)\|_p=\beta_p} \|G^n_{ii}(:,k)-G_{ii}(:,k)\|^2_2, \, \forall i \notag \\
\mbox{and}\qquad&\min_{\|G_{ij}(:,k)\|_p\leq \mu_p} \|G^n_{ij}(:,k)-G_{ij}(:,k)\|^2_2, \, \forall i\neq j \notag.
\end{align}
In other words for $p=1,\,2,\, \infty$ we need to solve problems of the form 
\begin{align}\label{eq:genprob}
\min_{\|g\|_p=\beta_p} \|g-h\|^2_2 \qquad \mbox{and} \qquad \min_{\|g\|_p\leq \mu_p} \|g-h\|^2_2.
\end{align}
The solutions are collected in the following Theorem.
\begin{Theorem} Denote by $g_{\beta_p}$ the minimal argument of the first problem and by $g_{\mu_p}$ the minimal argument of the second problem in~\eqref{eq:genprob}. \\
$p=1:\:$ Set $\sigma(i)=\sign(h(i))$ if $h(i)\neq0$ and $\sigma(i)=1$ else, and denote by $m$ the length of the $h$, then
\begin{align} 
g_{\beta_1}(i)= h(i) + \sigma(i) \lambda, \, \mbox{ where } \lambda= \frac{\beta-\|h\|_1}{m}.\notag
\end{align}
If $\|h\|_1\leq \mu$ set $g_{\mu_1}=h$. Otherwise set $g^0=h$ and iteratively shrink
\begin{align}
 g^k_{\beta_1}(i)= \sigma(i) \max(|g^{k-1}(i)| &- \lambda^k,0), \notag \\
 & \mbox{ where } \lambda^k= \frac{\|g^{k-1}\|_1-\mu}{\sharp \{g_i^{k-1}\neq 0\}}.\notag
\end{align}
until $g^k$ with $\|g^k\|_1=\mu$ is found and set  $g_{\mu_1}=g^k$.\\
\noindent $p=2:\:$
\begin{align} 
&g_{\beta_2}=\beta_2 \cdot \frac{h}{\|h\|_2},  \notag \\ 
&g_{\mu_2} = \left\{ \begin{array}{cl} % brackets may be (...), [...], \{...\}, or left out
      h & \mbox{ if } \|h\|_2 \leq \mu_2\\
     \mu_2 \cdot \frac{h}{\|h\|_2}& \mbox{ else} \\
   \end{array} \right. . \notag
\end{align}
\noindent $p=\infty:$ Let $ i_{\max} $ be the index of (one of) the largest absolute component of $h$ then
\begin{align} 
&g_{\beta_\infty}(i) = \left\{ \begin{array}{cl} % brackets may be (...), [...], \{...\}, or left out
     1 & \mbox{ if } i =  i_{\max}\\
     h(i) & \mbox{ else} \\
   \end{array} \right. ,
\notag \\
&g_{\mu_\infty}(i) = \left\{ \begin{array}{cl} % brackets may be (...), [...], \{...\}, or left out
      h(i) & \mbox{ if } |h(i)| \leq \mu_\infty\\
     \mu_\infty & \mbox{ else} \\
   \end{array} \right. . \notag
\end{align}

\end{Theorem}
Lastly as initialisation $G^0$ we choose the projection of $\hat{G}$ onto $\mathcal{G}_\mu$, \ie $G^0=\prox_{f_1}(\hat{G})$. Finding the correct step-sizes is usually a matter or trial and error. For the application considered here we used $\gamma^n= \|G^n\|_\mathbf{2}/(20 \|\nabla f_2(G^n)\|_\mathbf{2})$, which worked better for small $\mu$, and $\gamma^n= 1/ \|\nabla f_2(G^n)\|_\mathbf{2}$, which worked better for large $\mu$. The iteration was stopped when the relative improvement in each step was below $10^{-4}$. The number of iterations for the alternative projections was set to 10. \\

\noindent {\bf Thanks:} We would like to thank John Wright for helping us getting the cropped version of the AR-database faces.

\bibliographystyle{abbrv}
\bibliography{/Users/karinschnass/Desktop/latexnotes/karinbibtex}

\end{document}